\documentclass[letterpaper]{article} % DO NOT CHANGE THIS
\usepackage{aaai2026}  % DO NOT CHANGE THIS
\usepackage{times}  % DO NOT CHANGE THIS
\usepackage{helvet}  % DO NOT CHANGE THIS
\usepackage{courier}  % DO NOT CHANGE THIS
\usepackage[hyphens]{url}  % DO NOT CHANGE THIS
\usepackage{graphicx} % DO NOT CHANGE THIS
\usepackage{natbib}  % DO NOT CHANGE THIS AND DO NOT ADD ANY OPTIONS TO IT
\usepackage{caption} % DO NOT CHANGE THIS AND DO NOT ADD ANY OPTIONS TO IT
\usepackage{algorithm}
\usepackage{algorithmic}
\usepackage{amsmath,amsfonts,amssymb}
\usepackage{newfloat}
\usepackage{listings}
\DeclareCaptionStyle{ruled}{labelfont=normalfont,labelsep=colon,strut=off} % DO NOT CHANGE THIS
\floatstyle{ruled}
\newfloat{listing}{tb}{lst}{}
\floatname{listing}{Listing}
\usepackage{subcaption}
\usepackage{booktabs}
\usepackage{tabularx}
\usepackage{array}
\usepackage[table]{xcolor}

\usepackage{multirow, xcolor, colortbl}
\usepackage{cuted}
\usepackage{tabularx}
\usepackage{multirow}
\usepackage{multicol}
\usepackage{amsmath}
\usepackage{amsfonts}
\usepackage{booktabs}
\usepackage{array}

\title{DCAC: Dynamic Class-Aware Cache \\Creates Stronger Out-of-Distribution Detectors}
\author{
    Yanqi Wu\textsuperscript{\rm 1,2,3}, 
    Qichao Chen\textsuperscript{\rm 4}, 
    Runhe Lai\textsuperscript{\rm 1,2,3}, 
    Xinhua Lu\textsuperscript{\rm 1,2,3}, \\
    Jia-Xin Zhuang\textsuperscript{\rm 5}\thanks{Corresponding author.}, 
    Zhilin Zhao\textsuperscript{\rm 1,3}, 
    Wei-Shi Zheng\textsuperscript{\rm 1,2,3}, 
    Ruixuan Wang\textsuperscript{\rm 1,2,3*}
}
\affiliations{
    \textsuperscript{\rm 1}School of Computer Science and Engineering, Sun Yat-sen University, Guangzhou, China\\
    \textsuperscript{\rm 2}Peng Cheng Laboratory, Shenzhen, China \\
    \textsuperscript{\rm 3}Key Laboratory of Machine Intelligence and Advanced Computing, MOE, Guangzhou, China\\
    \textsuperscript{\rm 4}University of Nottingham Malaysia, Semenyih, Malaysia\\
    \textsuperscript{\rm 5}Hong Kong University of Science and Technology, Hong Kong, China \\
    
    wuyq268@mail2.sysu.edu.cn,
    hcxqc1@nottingham.my,
    \{lairh5,luxh55\}@mail2.sysu.edu.cn,
    jzhuangad@cse.ust.hk,
    zhaozhlin@mail.sysu.edu.cn,
    wszheng@ieee.org,
    wangruix5@mail.sysu.edu.cn
}

\usepackage{bibentry}
\begin{document}

\maketitle

\begin{abstract}
Out-of-distribution (OOD) detection remains a fundamental challenge for deep neural networks, particularly due to overconfident predictions on unseen OOD samples during testing. We reveal a key insight: OOD samples predicted as the same class, or given high probabilities for it, are visually more similar to each other than to the true in-distribution (ID) samples. Motivated by this class-specific observation, we propose DCAC (Dynamic Class-Aware Cache), a training-free, test-time calibration module that maintains separate caches for each ID class to collect high-entropy samples and calibrate the raw predictions of input samples. DCAC leverages cached visual features and predicted probabilities through a lightweight two-layer module to mitigate overconfident predictions on OOD samples. This module can be seamlessly integrated with various existing OOD detection methods across both unimodal and vision-language models while introducing minimal computational overhead. Extensive experiments on multiple OOD benchmarks demonstrate that DCAC significantly enhances existing methods, achieving substantial improvements, i.e., reducing FPR95 by 6.55\% when integrated with ASH-S on ImageNet OOD benchmark. 
\end{abstract}

\begin{links}
    \link{Code}{https://github.com/wyqstan/DCAC}
\end{links}
\section{Introduction}
In recent years, deep neural networks(DNNs) have demonstrated remarkable performance on many challenging tasks. However, their reliability drops significantly in open-world settings when encountering previously unseen classes, often producing overconfident predictions for OOD inputs. This poses significant safety risks in critical applications like autonomous driving and medical diagnosis. Existing methods typically fall into two groups: designing new scoring functions~\cite{hendrycks2016baseline,hendrycks2019scaling,liu2020energy,ming2022delving}, or training with auxiliary outliers~\cite{chen2024tagfog,hendrycks2018deep,wu2025pursuing}. The latter %, known as Outlier Exposure (OE), 
is often more effective due to the usage of explicit outlier knowledge.

\begin{figure}[t]
    \centering
    \includegraphics[width=\linewidth]{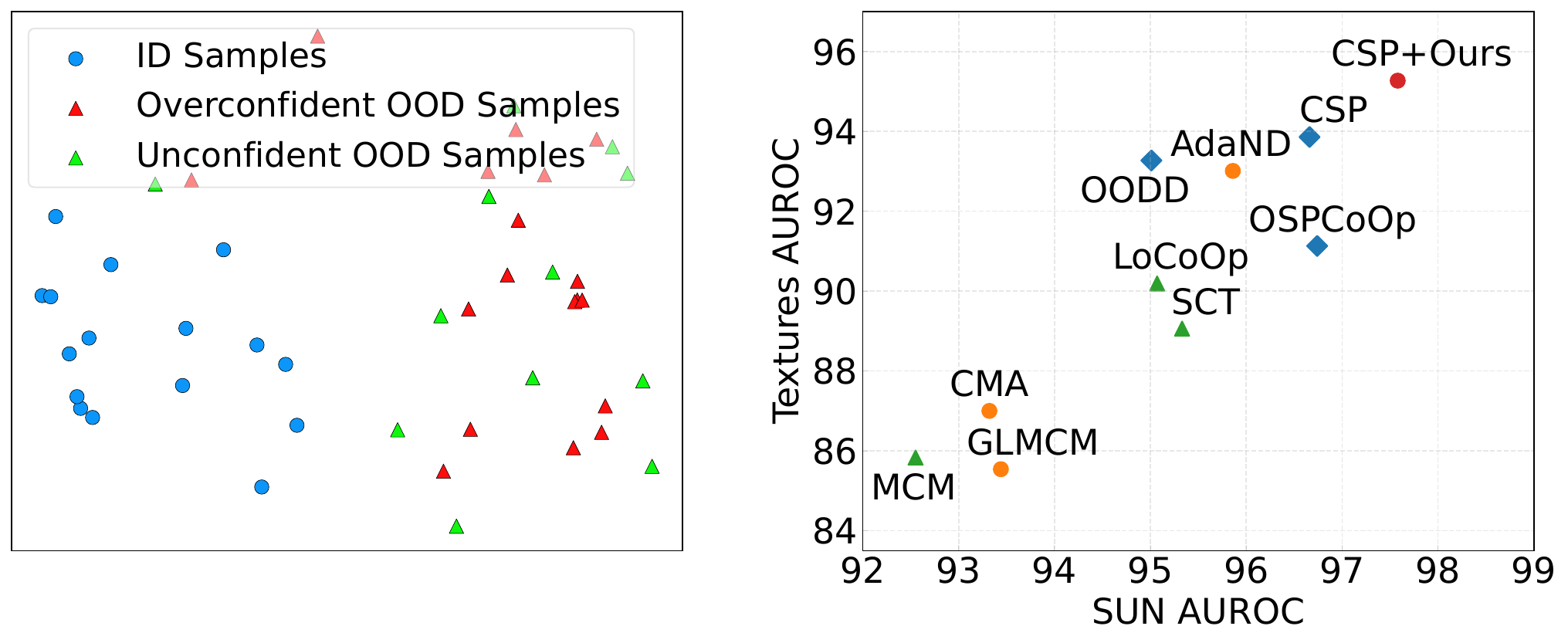}
    \caption{The t-SNE visualization of normalized image features for test samples predicted as the same class (left) and OOD detection performance using CLIP-B/16 (right).}
    \label{fig:feature_visualize}
\end{figure}

However, if the OOD samples encountered during testing differ greatly from the outliers seen during training, the model may still produce overconfident predictions on these unseen OOD inputs. To address this limitation, test-time OOD detection has emerged as a promising direction. Existing methods~\cite{fan2024test,yang2023auto,yang2025oodd} leverage test samples to perform immediate calibration or parameter updates to the model, improving robustness against previously unseen OOD inputs. Most of these methods rely on the predictive confidence, typically measured by the softmax probability assigned to the predicted class. However, they often overlook the rich information embedded in the predicted labels assigned to the OOD samples.
 This raises a natural question:
\begin{center}
    \textit{Do OOD samples sharing the same predicted labels exhibit any underlying relationships?}
\end{center}
To answer this question, we analyzed samples classified into the same class and uncovered an interesting phenomenon. As illustrated on the left side of Fig.~\ref{fig:feature_visualize}, OOD samples that are predicted as the same class exhibit significantly higher internal visual similarity. To further investigate this phenomenon, we categorize these OOD samples into overconfident and unconfident groups based on their predictive entropy, finding that visual similarity between unconfident and overconfident OOD samples tends to be higher than that between unconfident OOD and ID samples. In addition, we observe that some OOD samples, although not predicted as this class, are assigned high predicted probabilities for it and also share a relatively high degree of visual similarity with the aforementioned OOD samples predicted as this class.

Based on this class-specific observation, we propose  collecting unconfident OOD samples predicted as each ID class and using their visual similarity to overconfident OOD samples as a reference to calibrate the predictions of the overconfident ones. Motivated by this, we propose DCAC (Dynamic Class-Aware Cache), a training-free test-time calibration module. DCAC maintains separate caches for each ID class to collect unconfident samples and uses their visual features and predicted probabilities to calibrate overconfident predictions through a lightweight two-layer module inspired by Tip-Adapter~\cite{zhang2022tip}.  For each test sample, its visual features are processed through this module to generate calibration logits that are combined with original predictions for enhanced OOD detection.

DCAC offers several practical advantages: it is a plug-and-play module that is architecture-agnostic, requires no additional training, and seamlessly integrates with diverse OOD detection methods across both unimodal and vision-language models(VLMs), further enhancing their OOD detection performance. Comprehensive evaluations demonstrate that DCAC consistently enhances existing baselines with substantial performance gains. The main contributions of this work are summarized below. 
\begin{itemize}
\item We reveal that OOD samples predicted as the same class or assigned relatively high probabilities for that class tend to share higher visual similarity with each other than with true ID samples, and propose leveraging unconfident samples to calibrate overconfident OOD predictions.
\item We introduce DCAC, a training-free and architecture-agnostic module that calibrates predictions using the visual features and probabilities from test samples.
\item We show that DCAC can be flexibly fused  %broadly compatible 
with existing OOD detection methods, achieving consistent performance improvements across multiple benchmarks.
\end{itemize}
\section{Related Works}
\noindent\textbf{Traditional Unimodal OOD Detection.}
OOD detection has attracted growing research attention in recent years. Many methods have emerged to address the issue, which can generally be categorized into two groups: post-hoc methods and training-based methods.
Post-hoc methods typically design an OOD scoring function by analyzing model outputs such as logits, feature representations or layer-wise statistics.
MSP~\cite{hendrycks2016baseline} is the first baseline using the maximum softmax probability as the OOD score. ODIN~\cite{liang2017enhancing} enhances MSP by applying temperature scaling and input perturbation. Energy score~\cite{liu2020energy} is introduced as a new scoring function for OOD detection. ReAct~\cite{sun2021react} suppresses high activation values in the penultimate layer of visual features, while ASH~\cite{djurisic2023extremely} discards most activations in that layer. OptFs~\cite{zhao2024towards} optimizes feature shaping via a closed-form solution. CADRef~\cite{ling2025cadref} enhances feature discrimination by decoupling class-aware relative features based on sign alignment with model weights. On the other hand, training-based methods often use outlier knowledge during training. MOS~\cite{huang2021mos} partitions ID classes into semantically distinct groups and treats samples from other groups as OOD during training. TagFog~\cite{chen2024tagfog} creates OOD samples by shuffling ID image patches, and FodFom~\cite{chen2024fodfom} utilizes Stable Diffusion to synthesize training OOD samples that closely resemble ID images. % to aid training. 

\noindent\textbf{VLM-based OOD Detection.} With rapid development of VLMs such as CLIP~\cite{radford2021learning}, more and more researchers have begun to focus on the OOD detection problem in VLMs. MCM~\cite{ming2022delving} was the first method to adapt CLIP for OOD detection by combining MSP with temperature scaling. GLMCM~\cite{miyai2023zero} further incorporates the maximum probability of local image patches. NegLabel ~\cite{jiang2024negative}, CSP~\cite{chen2024conjugated} and CMA~\cite{lee2025concept} introduce text information that is semantically unrelated to ID classes to improve OOD discrimination. These are post-hoc methods built on vanilla CLIP. Many recent methods apply finetuning to adapt CLIP to downstream tasks. LoCoOp~\cite{miyai2023locoop} and SCT~\cite{yu2024self} use entropy maximization strategies to push OOD visual embeddings away from ID-aligned text features in the multimodal space. Local-Prompt~\cite{zeng2025local} freezes the global prompt while learning local prompts with region-level regularization and negative sample enhancement.  OspCoOp~\cite{xu2025ospcoop} leverages ID background as auxiliary training OOD data. % during fine-tuning. 

\begin{figure*}[t]
\centering
\includegraphics[width=\textwidth]{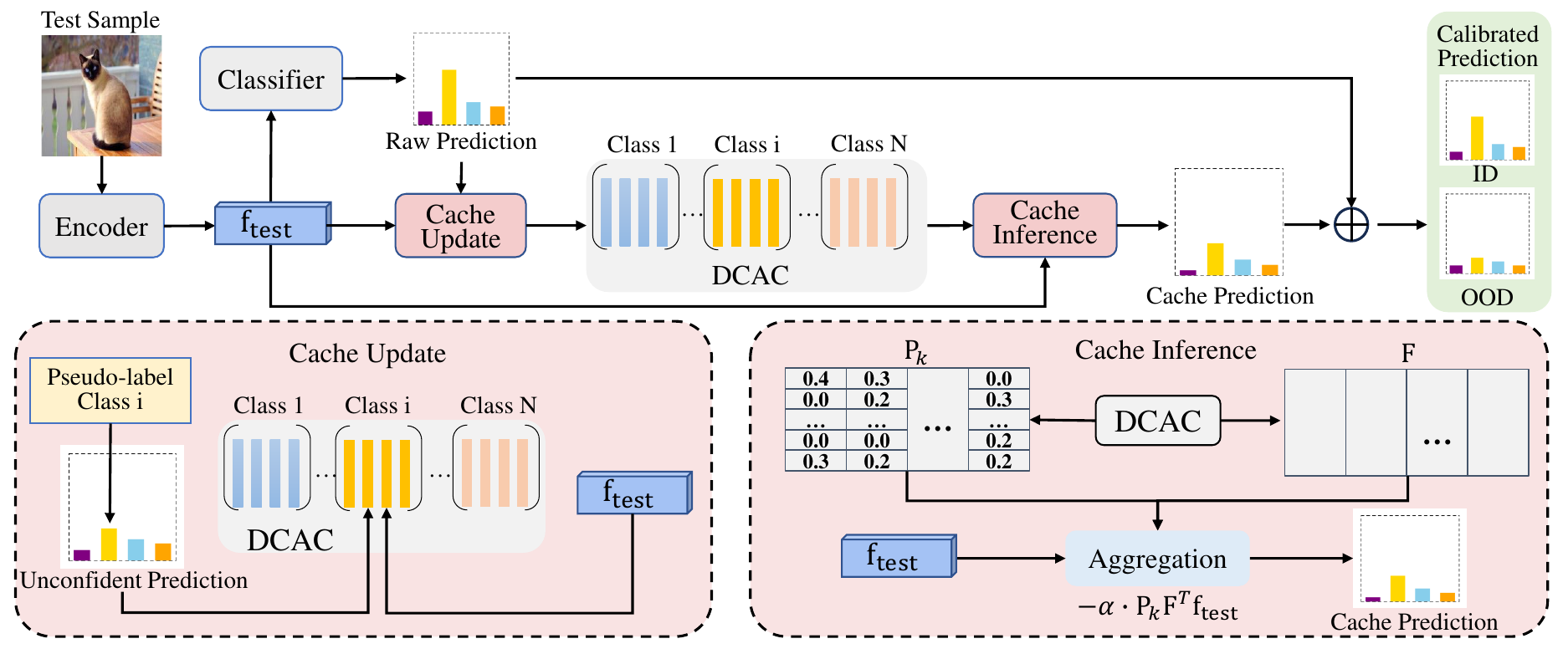} % Reduce the figure size so that it is slightly narrower than the column.
\caption{Overview of the Dynamic Class-aware Cache (DCAC) framework. DCAC maintains caches for each ID class to collect unconfident samples during testing guided by their entropy and the cache capacity. For calibration, cache samples generate a signal for each test sample which is combined with raw prediction to produce calibrated outputs for OOD detection.}
\label{fig:framework}
\end{figure*}
\noindent\textbf{Test-Time OOD Detection.} Very recently, increasing attention has been given to leveraging information from real test-time scenarios to assist OOD detection. AUTO~\cite{yang2023auto} performs test-time adaptation by applying stochastic gradient descent directly to the model, aiming to reduce confidence on potential OOD samples. Unlike AUTO that update the entire model that suffers from performance degradation under noisy test streams, AdaND~\cite{cao2025noisy} freezes the model and just trains a lightweight noise detector to filter out OOD samples. RTL++~\cite{fan2024test} assumes a linear correlation between OOD scores and features. However, this simple assumption is challenged by the overlap in feature distributions between hard OOD and ID samples. TULIP~\cite{zhang2025tulip} introduces priors of ID and OOD distributions at test time to calibrate model uncertainty.
Our proposed method calibrates model prediction based on the image features and the raw predictions of test samples , without relying on any external prior. OODD~\cite{yang2025oodd} maintains a priority queue to accumulate representative OOD features and leveraging them to calibrate the OOD score of test samples. Unlike OODD that updates a queue solely based on OOD scores, this study designs and updates the cache based on the entropy and predicted label of each test sample. Furthermore, our method focuses on calibrating output logits, whereas OODD adjusts the final OOD scores.

\section{Methodology}

\subsection{Preliminaries}
\textbf{OOD Detection.} 
Consider a $C$-class classification problem where a deep neural network is trained on an ID training dataset $\mathcal{D}_\text{train}$. For an input image $\mathbf{x}$, the network produces a feature vector $\mathbf{f} = \mathcal{F}(\mathbf{x}) \in \mathbb{R}^{d}$ through feature extractor $\mathcal{F}$, and class logits $\mathbf{z}$ through a linear classifier: 
\begin{equation}
\mathbf{z} = \mathbf{W}^\top \mathbf{f} + \mathbf{b} \,,
\end{equation}
where $\mathbf{W} \in \mathbb{R}^{d \times C}$ and $\mathbf{b} \in \mathbb{R}^C$ are learnable parameters.

OOD detection determines whether the input $\mathbf{x}$ belongs to the training distribution. A scoring function $\mathcal{S}(\mathbf{z})$ is designed often based on the logits to assess distributional membership. Given threshold $\tau$, the detector outputs:
\begin{equation}
\label{eq:detector}
\mathcal{G}(\mathbf{x}) =
\begin{cases}
\text{ID}, & \text{if } \mathcal{S}(\mathbf{z}) \geq \tau \\
\text{OOD}, & \text{otherwise}
\end{cases}.
\end{equation}
\subsection{Overview}
We propose a test-time prediction calibration method (Fig.~\ref{fig:framework}) based on a key observation: for samples predicted as the same class, unconfident OOD samples show greater visual similarity to overconfident OOD samples than to true ID samples within that class. Our approach leverages this property by collecting unconfident samples during testing to calibrate overconfident predictions. The method has two components. First, we maintain dynamic class-aware caches that collect high-entropy samples and adapt to distribution shifts (Section 3.3). Second, we calibrate predictions using cached features and probability distributions as adaptive weights (Section 3.4). The calibrated predictions are then processed by OOD detection scoring functions $\mathcal{S}(\cdot)$. %, as detailed in Figure 2.

\subsection{Dynamic Class-Aware Caches}
 To calibrate predictions of overconfident OOD samples, we first need to collect unconfident samples with similar characteristics. We maintain a separate cache for each ID class, initially as an empty set, that collects high-entropy samples during testing. Higher entropy indicates higher prediction uncertainty, making these samples likely candidates for OOD data that can inform calibration. We implement an entropy-based gating mechanism that admits only samples whose entropy exceeds threshold $\delta$. Since the true OOD entropy distribution is unknown at test time, we estimate $\delta$ as the $\beta$-th percentile of the entropy distribution from ID training data, with $\beta=95$ in our experiments. This method ensures we preferentially collect potential OOD samples over confident ID samples.
 
 To maintain computational efficiency, each cache has fixed capacity and follows a first-in-first-out (FIFO) update policy. When a high-entropy sample arrives, we store its visual features $\mathbf{f}$ and  predicted probability distribution $\mathbf{p}$ in the cache for its predicted class. If the cache is full, we remove the oldest sample before storing a new one. This dynamic updating captures recent distribution patterns and enables rapid adaptation to evolving test conditions, providing fresh knowledge for calibrating subsequent predictions. 

\subsection{Calibrate Model Outputs}
Whenever a test sample is processed by the model and the cache has been updated accordingly, we immediately utilize all the information currently stored in the cache to calibrate the raw predictions of this sample. Let $\mathbf{F} = [\mathbf{f}_1, \mathbf{f}_2, ..., \mathbf{f}_N] \in \mathbb{R}^{d \times N}$ denote the cached feature matrix and  $\mathbf{P} = [\mathbf{p}_1, \mathbf{p}_2, ..., \mathbf{p}_N] \in \mathbb{R}^{C \times N}$ denote the corresponding probability matrix, where $d$ is the visual feature dimension, $N$ is the number of cached samples across all classes, and $C$ is the number of ID classes. Based on $\mathbf{F}$ and $\mathbf{P}$, a novel two-layer linear transformation module is designed. Specifically, the first layer is fully connected with weight parameters $\mathbf{F}$. %is used as the weight parameters of the first layer, 
This layer can be interpreted as computing the cosine similarity between the visual feature of the test sample and those of the cached samples. Similarly, the second layer is also fully connected with the weight parameters $\mathbf{P}$, which functions as a weighted aggregation on the similarity scores obtained from the first layer. Note that the weight parameters of the two-layer module are not learned, but set directly.

To reduce the influence of less relevant predictions from the cached samples, we apply a sparsification to the cached probability matrix $\mathbf{P}$. Specifically, for each cached sample, we retain only the top-$k$ largest values in its probability distribution and set the rest entries to zero. This operation can be formulated as:
\begin{equation}
\mathbf{P}_{k} = \mathbf{P} \odot \mathbf{M},
\end{equation}
where $\mathbf{M} \in \{0, 1\}^{C \times N}$ is a binary mask that retains only the top-$k$ entries in each column of $\mathbf{P}$, and $\odot$ denotes element-wise multiplication.

Ultimately, the cache prediction vector, which shares the same dimension as the original logit output, is computed as:
\begin{equation}
% \small
\mathbf{z}_{\mathrm{cache}} = -\mathbf{P}_{k} \mathbf{F}^\intercal
\mathbf{f}_{\mathrm{test}} ,
\end{equation}
where $\mathbf{f}_{\text{test}} \in \mathbb{R}^d$ denotes the visual feature of the test sample. We take the negative of $\mathbf{z}_{\mathrm{cache}}$ before combining it with the raw prediction $\mathbf{z}$. Since overconfident OOD samples yield high values for the predicted class in both $\mathbf{z}$ and $\mathbf{z}_{\mathrm{cache}}$, introducing the negative sign helps suppress excessive confidence in the final calibrated prediction. 
The original logit output is finally calibrated as: 
\begin{equation}
% \small
\hat{\mathbf{z}} = \mathbf{z} + \alpha \cdot \mathbf{z}_{\mathrm{cache}},
\end{equation}
where $\alpha$ is a hyperparameter controlling calibration strength. $\hat{\mathbf{z}}$ replaces $\mathbf{z}$ in Eq.~\ref{eq:detector} to be fed into various scoring functions for OOD detection. In particular, the prediction of some recent methods~\cite{chen2024conjugated,jiang2024negative} includes both ID and OOD classes, our method only calibrates the ID-relevant components. The complete algorithmic process is detailed in Algorithm~\ref{alg:ood_calibration}.

\begin{algorithm}[!t]
\small
\caption{Test-time Calibration for a Single Sample}
\label{alg:ood_calibration}
\begin{algorithmic}[1]
\REQUIRE Test sample $\mathbf{x}$, model $f$, cache $\mathcal{C}$, entropy threshold $\delta$, top-$k$, correction factor $\alpha$, max cache size $m$
\STATE $\mathbf{z}, \mathbf{f}_{\mathrm{test}} \leftarrow f(\mathbf{x})$
\STATE $\mathbf{p} \leftarrow \mathrm{softmax}(\mathbf{z}),\quad H \leftarrow H(\mathbf{p}),\quad \hat{y} \leftarrow \arg\max(\mathbf{p})$
\IF{$H > \delta$}
    \STATE $\mathcal{B} \leftarrow \mathcal{C}[\hat{y}]$ \hfill \COMMENT{Select buffer for predicted class}
    \IF{$|\mathcal{B}| \geq m$}
        \STATE Remove the oldest element from $\mathcal{B}$
    \ENDIF
    \STATE Append $(\mathbf{f}_{\mathrm{test}}, \mathbf{p})$ to $\mathcal{B}$
\ENDIF

\STATE \textbf{Calibration:}
\STATE $\mathbf{F} \leftarrow [\mathbf{f}_1, \mathbf{f}_2, ..., \mathbf{f}_N] \in \mathbb{R}^{d \times N}$ 
\STATE $\mathbf{P} \leftarrow [\mathbf{p}_1, \mathbf{p}_2, ..., \mathbf{p}_N] \in \mathbb{R}^{C \times N}$

\STATE Construct mask $\mathbf{M} \in \{0,1\}^{C \times N}$ such that only top-$k$ entries in each column of $\mathbf{P}$ are 1
\STATE $\mathbf{P}_{k} \leftarrow \mathbf{P} \odot \mathbf{M}$ \hfill \COMMENT{Element-wise sparsification}
\STATE $\mathbf{z}_{\mathrm{cache}} \leftarrow -\mathbf{P}_{k} \mathbf{F}^\intercal
\mathbf{f}_{\mathrm{test}}$

\STATE $\hat{\mathbf{z}} \leftarrow \mathbf{z} + \alpha \cdot \mathbf{z}_{\mathrm{cache}}$
\RETURN $\hat{\mathbf{z}}$
\ENSURE $\hat{\mathbf{z}}$
\end{algorithmic}
\end{algorithm}

\begin{table*}[!ht]
\centering
\small
\begin{tabular}{clccccccccccccccc}
\toprule
\multirow{3}{*}{\begin{tabular}[c]{@{}c@{}}{Architecture}\end{tabular}} 
&\multirow{3}{*}{Method} & \multicolumn{8}{c}{OOD Datasets} & \multicolumn{2}{c}{\multirow{2}{*}{Average}} \\ \cline{3-10} 
&  & \multicolumn{2}{c}{iNaturalist} & \multicolumn{2}{c}{SUN} & \multicolumn{2}{c}{Textures} & \multicolumn{2}{c}{Places} \\
\cmidrule(lr){3-4} \cmidrule(lr){5-6} \cmidrule(lr){7-8} \cmidrule(lr){9-10} \cmidrule(lr){11-12}
& & F$\downarrow$ & A$\uparrow$ 
& F$\downarrow$ & A$\uparrow$ 
& F$\downarrow$ & A$\uparrow$ 
& F$\downarrow$ & A$\uparrow$ 
& F$\downarrow$ & A$\uparrow$ \\ 
\midrule
\multirow{8}{*}{\begin{tabular}[c]{@{}c@{}}ResNet50\end{tabular}} 
    & MSP & 52.73 & 88.42 & 68.58 & 81.75 & 66.15 & 80.46 & 71.59 & 80.63 & 64.76 & 82.82 \\
    & Energy  & 53.96 & 90.59 & 58.28 & 86.73 & 52.3 & 86.73 & 65.43 & 84.12 & 57.49 & 87.04 \\
    & ReAct  & 19.55 & 96.39 & 24.01 & 94.41 & 45.83 & 90.45 & 33.45 & 91.93 & 30.71 & 93.30 \\
    & OptFS & 16.79 & 96.88 & 35.31 & 93.13 & 23.08 & 95.74 & 44.78 & 90.42 & 29.99 & 94.04 \\
   & CADRef & 16.08 & 96.9 & 39.23 & 91.26 & 12.60 & 97.14 & 51.12 & 87.8 & 29.76 & 93.28 \\
    & OODD & \underline{7.13} & \textbf{98.71} & 42.36 & 92.07 & 16.16 & 97.01 & 53.59 & 87.13 & 29.81 & 93.73 \\
    & ASH-S  & 11.49 & 97.87 & 27.96 & 94.02 & \underline{11.97} & \underline{97.60} & 39.83 & 90.98 & 22.81 & 95.12 \\
    & LINE & 12.26 & 97.56 & 19.48 & 95.26 & 22.54 & 94.44 & \textbf{28.52} & 92.85 & 20.70 & 95.03 \\
    & DDCS & 11.63 & 97.85 & \underline{18.63} & \underline{95.68} & 18.40 & 95.77 & \underline{28.78} & \underline{92.89} & \underline{19.36} & \underline{95.55} \\
    & \textbf{ASH-S+Ours} & \textbf{6.69} & \underline{98.54} & \textbf{17.47} & \textbf{96.21} & \textbf{11.70} & \textbf{97.74} & 29.18 & \textbf{93.11} & \textbf{16.26} & \textbf{96.40} \\

 \midrule
 \multirow{8}{*}{\begin{tabular}[c]{@{}c@{}}CLIP-B/16\end{tabular}} 
    & MCM & 31.95 & 94.16 & 37.22 & 92.55 & 58.35 & 85.83 & 42.98 & 90.10 & 42.63 & 90.66 \\
    & SCT & 13.94 & 95.86 & 20.55 & 95.33 & 41.51 & 89.06 & 29.86 & 92.24 & 26.47 & 93.37 \\
    & CMA  & 8.88 & 98.19 & 29.03 & 93.32 & 51.26 & 87.00 & 27.60 & 93.87 & 29.19 & 93.10 \\
    & AdaND & 4.19 & 98.91 & 17.08 & 95.86 & \underline{21.76} & 93.01 & \textbf{20.95} & \textbf{94.55} & \underline{16.00} & 95.58 \\
    & OODD  & 2.22 & 99.36 & 21.49 & 95.01 & 30.69 & 93.27 & 44.76 & 87.10 & 24.79 & 93.69 \\  
    & OSPCoOp & 15.25 & 97.13 & 18.26 & 96.74 & 41.26 & 91.13 & 25.74 & \underline{94.01} & 25.13 & 94.75 \\
    & CSP & \underline{1.54} & \underline{99.60} & \underline{13.66} & \underline{96.66} & 25.52 & \underline{93.86} & 29.32 & 92.90 & 17.51 & \underline{95.76} \\
    & \textbf{CSP+Ours} & \textbf{1.20} & \textbf{99.62} & \textbf{10.18} & \textbf{97.58} & \textbf{20.62} & \textbf{95.27} & \underline{25.29} & 93.94 & \textbf{14.32} & \textbf{96.60} \\
\bottomrule
\end{tabular}%
\caption{Comparison with state-of-the-art methods on the far-OOD benchmarks. The best results are in bold, while the second are underlined. $\uparrow$ indicates larger values are better and $\downarrow$ indicates smaller values are better.}
\label{tab:far}
\end{table*}

\section{Experiments}

\subsection{Experimental Setup}
\paragraph{Datasets.}
Following prior work~\cite{huang2021mos}, we use ImageNet-1K as ID dataset and evaluate on six OOD datasets: four far-OOD datasets including iNaturalist~\cite{van2018inaturalist}, SUN~\cite{xiao2010sun}, Places~\cite{zhou2017places}, and Textures~\cite{cimpoi2014describing}, and two near-OOD datasets, NINCO~\cite{bitterwolf2023ninco} and SSB-hard~\cite{vaze2022openset}. We further validate our method on five additional ID datasets: ImageNet100~\cite{ming2022delving}, UCF101~\cite{soomro2012ucf101}, FGVC-Aircraft~\cite{maji2013fine}, CIFAR10 and CIFAR100~\cite{krizhevsky2009learning}.

\noindent\textbf{Implementation Details.} 
Our method is implemented in PyTorch and all experiments are conducted on NVIDIA GeForce A30 GPUs with a batch size of 512. To simulate a realistic deployment scenario, all experiments are conducted by mixing ID and OOD data together and shuffling them. For each experiment, we use five different random seeds for shuffling. For the ImageNet-1K benchmark, we set the cache size per class $m$ to 20, calibration strength $\alpha$ to 0.9, and top $k$ to 20. %predicted probabilities for weighted averaging. 
Test samples with entropy above the 95th percentile ($\beta$ = 95) of ID training set are selected for caching.

\noindent\textbf{Metric.} The evaluation metrics include the false positive rate (F, FPR95) of OOD samples when the true positive rate of ID samples is at 95\%, and the area under the receiver operating characteristic curve (A, AUROC).

\noindent\textbf{Baseline.} 
We choose various competitive methods as baseline for comparison. For unimodal methods, we select MSP~\cite{hendrycks2016baseline}, Energy~\cite{liu2020energy}, ReAct~\cite{sun2021react}, ASH-S~\cite{djurisic2023extremely}, DDCS~\cite{ddcs}, LINE~\cite{line}, OODD~\cite{yang2025oodd} as baseline methods. For VLM-based method, we select  MCM~\cite{ming2022delving}, CSP~\cite{chen2024conjugated}, CMA~\cite{lee2025concept}, AdaND~\cite{cao2025noisy}, OODD, OSPCoOp~\cite{xu2025ospcoop} as baseline methods. Specifically, OODD and AdaND are test-time methods and AdaND achieves state-of-art performance on ImageNet-1K far-OOD benchmark. To ensure fair comparison, we retain the original hyperparameter settings of the baseline methods. 
% Our proposed DCAC can be integrated with a wide range of existing OOD detection methods. For single-modal methods, we integrate DCAC with MSP \cite{hendrycks2016baseline}, ODIN \cite{liang2017enhancing}, Energy \cite{liu2020energy}, ReAct \cite{sun2021react}, DICE \cite{sun2022dice}, ASH-S \cite{djurisic2023extremely}, OptFs \cite{zhao2024towards}, and CADRef \cite{ling2025cadref}, all based on ResNet-50 as the backbone. For VLM-based methods, we integrate DCAC with MCM \cite{ming2022delving}, CoOp\cite{zhou2021learning}, LoCoOp \cite{miyai2023locoop}, SCT \cite{yu2024self}, CSP \cite{chen2024conjugated}, Local-Prompt \cite{zeng2025local}, OspCoOp \cite{xu2025ospcoop}, using CLIP-B/16 as the backbone.  For test-time OOD detection method, we integrate with OODD \cite{yang2025oodd} with MCM as Score function. To ensure fair comparison, we retain the original hyperparameter settings of the baseline methods. 

\subsection{Performance on Benchmarks}
\begin{figure}[t]
    \centering
    \includegraphics[width=\columnwidth]{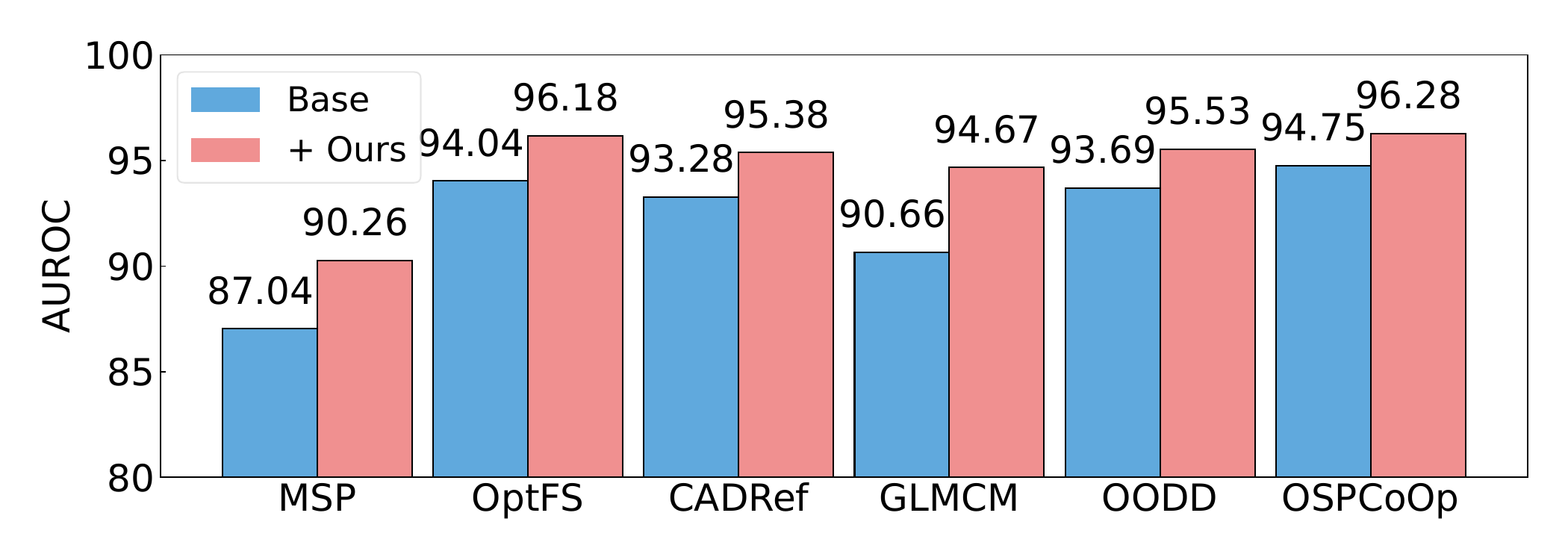}
    \caption{DCAC integrated with existing methods. }
    \label{fig:more_baseline}
\end{figure}

 \noindent\textbf{ImageNet-1K benchmarks.} As shown in Tab.~\ref{tab:far}, we compare the performance on the far-OOD benchmark using ResNet-50 and CLIP-B/16 as backbones. When combined with ASH-S and CSP, DCAC achieves superior performance, demonstrating the effectiveness of our method. Additional results of DCAC integrated with other OOD detection methods are illustrated in Fig.~\ref{fig:more_baseline}. Besides, we conduct compatibility experiments on the near-OOD benchmark, combining DCAC with various OOD detection methods (Tab.~\ref{table:comparison_hard}). All results support efficacy of our method.
 
 \noindent\textbf{Other ID benchmarks.} DCAC also provides effective prediction calibration on other ID datasets, as shown in Tab.~\ref{table:comparison_other_ID}. For the ImageNet100, UCF101 and FGVC-Aircraft datasets, their OOD datasets are the same as the far-OOD datasets used for ImageNet-1K. For the CIFAR datasets, their OOD datasets include SVHN~\cite{SVHN}, LSUN-R~\cite{LSUN}, LSUN-C, iSUN~\cite{iSUN}, Textures~\cite{cimpoi2014describing} and Places365~\cite{zhou2017places}. %More experiment details are provided in Appendix~\ref{detail_exp_results}. 

  % Tables \ref{tab:far} demonstrate the performance of DCAC combined with traditional single-modal and CLIP-based methods on far-OOD datasets. Our proposed method significantly improves performance across different OOD detection methods. Notably, when combined with the state-of-the-art CSP method, the average performance on far-OOD datasets improves considerably—FPR95 is reduced from 17.51\% to 14.32\%, and AUROC increases from 95.76\% to 96.60\%. Additionally, integrating with test-time detection methods like OODD further enhances performance, confirming the strong compatibility of our method.

\begin{table}[ht]
\small
\centering
\setlength{\tabcolsep}{1mm} % 控制列间距
\begin{tabular}{@{}llcccc@{}} \toprule
    \multirow{2.5}{*}{Architecture}  
    & \multirow{2.5}{*}{Method} 
    & \multicolumn{2}{c}{NINCO} 
    & \multicolumn{2}{c}{ssb-hard} \\ 
    \cmidrule(lr){3-4} \cmidrule(lr){5-6}
    & 
    & F$\downarrow$ & A$\uparrow$  
    & F$\downarrow$ & A$\uparrow$ \\ 
    \midrule
\multirow{7}{*}{ResNet50} 
    & MSP & 75.94 & 79.97 & 84.54 & 72.16 \\
    & \textbf{MSP+Ours} & \textbf{70.58} & \textbf{82.61} & \textbf{74.49} & \textbf{77.38} \\
    \cmidrule{2-6}
    & CADRef & 64.89 & 85.36 & 78.79 & 74.58 \\
    & \textbf{CADRef+Ours} & \textbf{63.64} & \textbf{85.46} & \textbf{74.32} & \textbf{78.71} \\
    \cmidrule{2-6}
    & ASH-S & 65.02 & 82.77 & 82.53 & 70.49 \\
    & \textbf{ASH-S+Ours} & \textbf{63.36} & \textbf{83.89} & \textbf{69.95} & \textbf{81.32} \\
\midrule
\multirow{7}{*}{CLIP/B-16} 
    & MCM & 79.50 & 74.57 & 89.90 & 62.95 \\
    & \textbf{MCM+Ours} & \textbf{75.34} & \textbf{75.72} & \textbf{83.40} & \textbf{69.41} \\
    \cmidrule{2-6}
    & LoCoOp & 78.27 & 73.58 & 87.92 & 64.10 \\
    & \textbf{LoCoOp+Ours} & \textbf{73.82} & \textbf{74.53} & \textbf{81.82} & \textbf{66.59} \\
    \cmidrule{2-6}
    & CSP & 69.01 & 77.69 & 82.74 & 71.93 \\
    & \textbf{CSP+Ours} & \textbf{66.71} & \textbf{80.52} & \textbf{78.90} & \textbf{76.12} \\
\bottomrule
\end{tabular}
\caption{OOD detection performance on near-OOD datasets.}
\label{table:comparison_hard}
\end{table}
\begin{table}[ht]
\centering
\small
\begin{tabular}{lccccc}
\toprule
\multirow{2.5}{*}{ID Dataset}  
 & \multicolumn{2}{c}{MCM} & \multicolumn{2}{c}{MCM+Ours} \\
\cmidrule(lr){2-3} \cmidrule(lr){4-5}
 & F$\downarrow$ & A$\uparrow$ & F$\downarrow$ & A$\uparrow$ \\
\midrule
ImageNet100   & 30.05 & 95.02 & \textbf{11.79} & \textbf{97.67} \\
UCF101        & 23.28 & 95.01 & \textbf{18.19} & \textbf{95.99} \\
FGVC-Aircraft & 13.92 & 96.70 & \textbf{11.19} & \textbf{97.41} \\
CIFAR10       & 15.09 & 96.02 & \textbf{12.73} & \textbf{96.51} \\
CIFAR100      & 86.11 & 76.42 & \textbf{68.93} & \textbf{79.76} \\
\bottomrule
\end{tabular}
\caption{OOD detection performance on other ID datasets.}
\label{table:comparison_other_ID}
\end{table}

% \paragraph{Detection performance across different model architectures.}
%  To verify the robustness and generality of our method, we integrate DCAC with various model architectures. As shown in Table \ref{table:comparison_other_arch}, we combine our method with the MSP method using ResNet50, DenseNet201, and ViT-B/16 backbones, as well as with the MCM method using CLIP models based on ResNet50, ViT-B/16, and ViT-L/14. Experimental results show consistent performance gains across all architectures, suggesting that our method is architecture-agnostic.

% \paragraph{Results on multiple ID datasets.}
%  In addition to using ImageNet-1K as the ID dataset, we also evaluate our method with ImageNet100, UCF101, FGVC-Aircraft, CIFAR10 and CIFAR100 as ID datasets. For the first three datasets, we use iNaturalist, SUN, Places, and Textures as OOD datasets. For the CIFAR datasets, we use SVHN, LSUN-R, LSUN-C, iSUN, Textures and Places365 as OOD datasets. As shown in Table \ref{table:comparison_other_ID}, our method continues to improve OOD detection performance on these ID dataset, reinforcing its generalizability.

\subsection{Analysis of the Proposed Method}
 For consistency, all subsequent experiments are conducted using the combination of DCAC and MCM as a representative example for in-depth analysis.
% \begin{figure}[!h]
%     \centering
%     \includegraphics[width=0.45\textwidth , height=0.2\textheight]{AnonymousSubmission/LaTeX/images/clip_auc_initialize.pdf}
%     \caption{AUROC comparison using different types of initialization (C-Out, T-Out, D-Out, Empty).}
%     \label{fig:auc_initial}
% \end{figure}
\begin{figure*}[!ht]
\centering
\includegraphics[width=\textwidth]{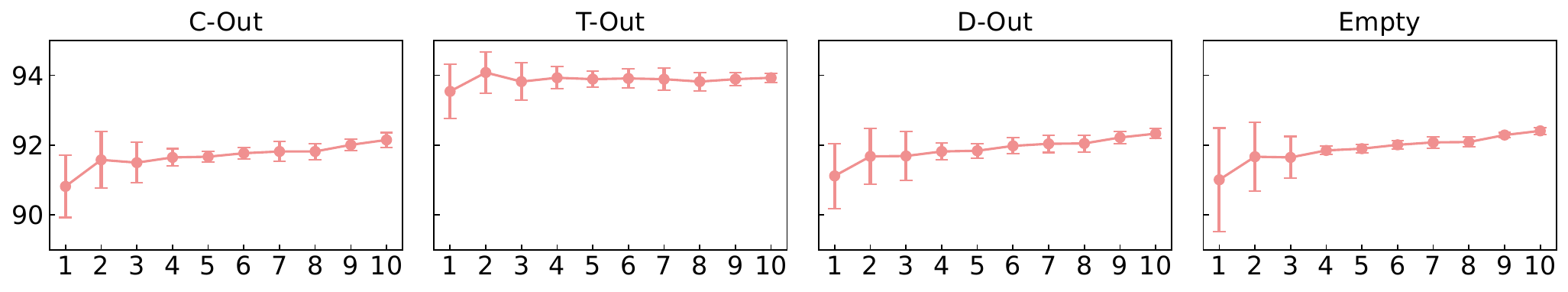} % Reduce the figure size so that it is slightly narrower than the column.
\caption{AUROC over the first 10 batches under different cache initialization strategies with mean and standard deviation.}
\label{fig:initial_places}
\end{figure*}

\begin{table}[ht]
\centering
% \small
% \setlength{\tabcolsep}{1mm}
\begin{tabular}{ccccc}
\toprule
Strategy & C-Out & D-Out & T-Out & Empty \\
\midrule
F$\downarrow$ & 25.76 & 24.73 & \textbf{21.53} & 24.25 \\
A$\uparrow$   & 94.35 & 94.60 & \textbf{95.28} & 94.68 \\
\bottomrule
\end{tabular}
\caption{OOD detection performance on ImageNet far-OOD benchmark with different cache initialized strategies.}
\label{tab:initialized}
\end{table}

\begin{figure}[t]
    \centering
    \includegraphics[width=\linewidth]{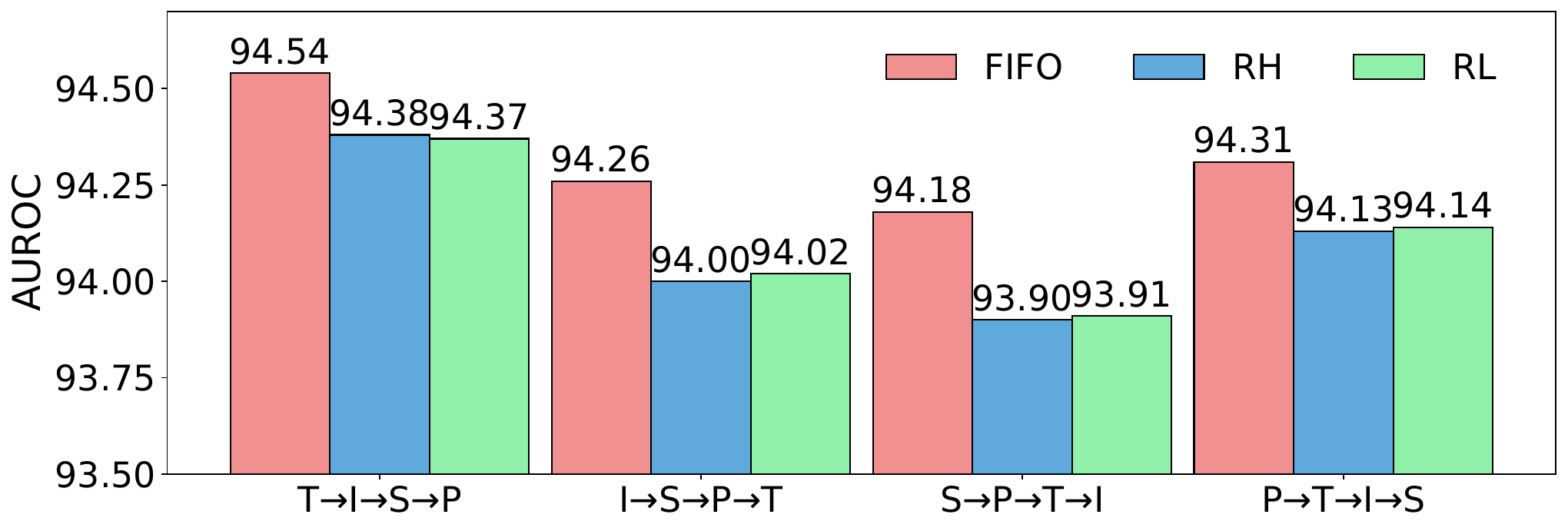}
    \caption{AUROC for the temporal drift OOD scenario settings with different update strategies. }
    \label{fig:update}
\end{figure}

\begin{figure}[t]
    \centering
    \includegraphics[width=\linewidth]{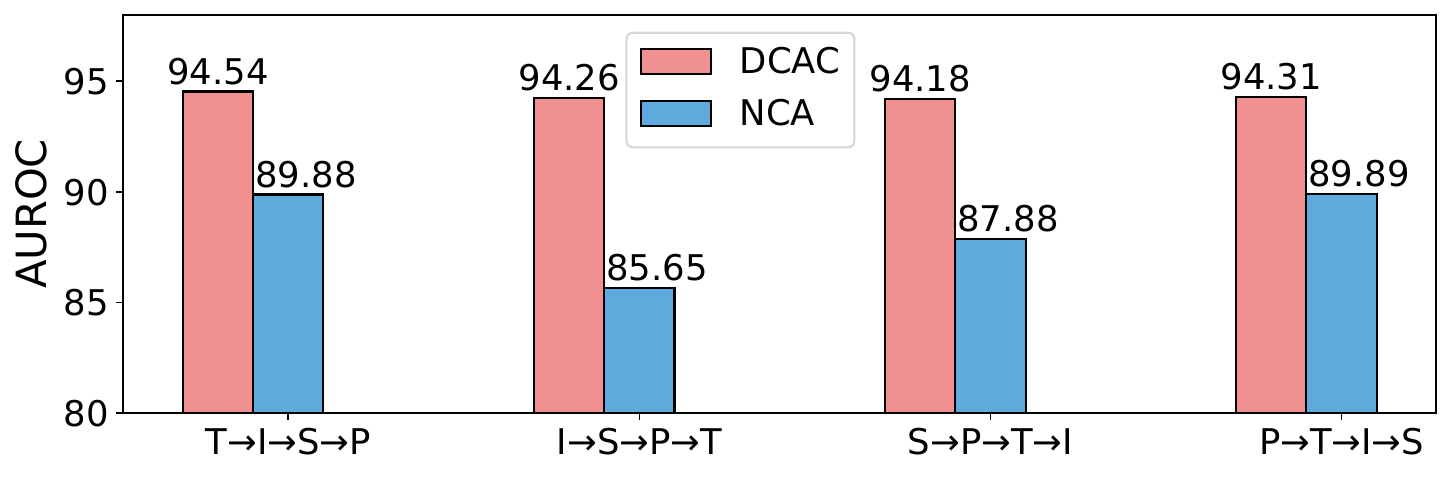}
    \caption{AUROC for the temporal drift OOD scenario settings with different cache construction strategies. }
    \label{fig:construction}
\end{figure}

\begin{figure*}[t]
\centering
\includegraphics[width=\textwidth]{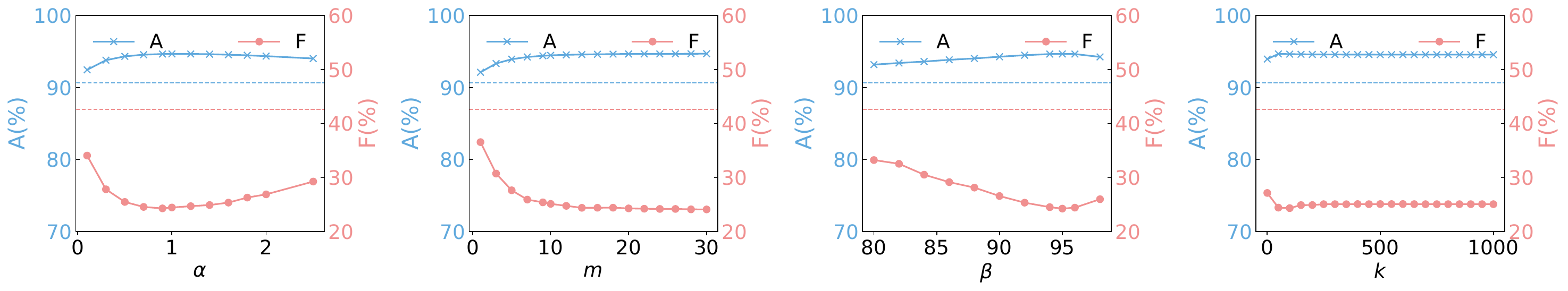} % Reduce the figure size so that it is slightly narrower than the column.
\caption{Sensitivity analysis of hyper-parameters. Each dashed line represents the baseline (MCM) performance.}
\label{fig:sensitivity}
\end{figure*}
\begin{figure}[!ht]
    \centering
    \includegraphics[width=\linewidth]{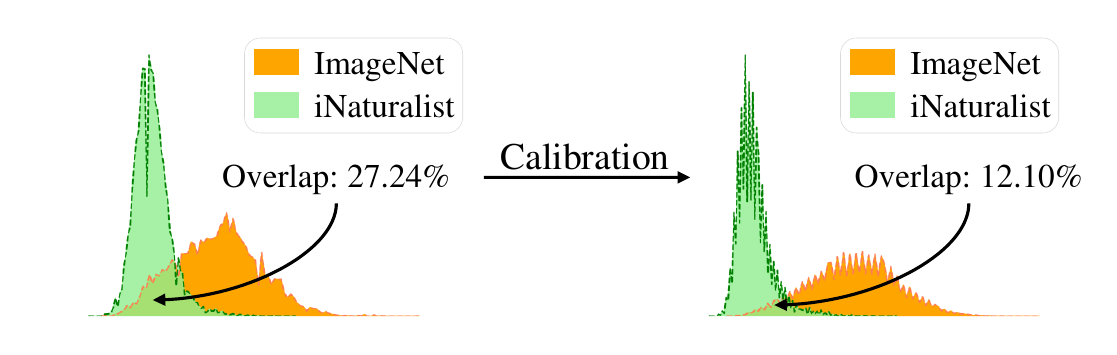}
    \caption{Comparison of the MCM score distribution on iNaturalist before and after calibration. }
    \label{fig:score_distribution}
\end{figure}
\noindent\textbf{Impact of Cache Initialization.}
Following OODD~\cite{yang2025oodd}, we present the results on the ImageNet-1K benchmark using DCAC with different cache initialization strategies in Tab.~\ref{tab:initialized}, including (1) Empty: initializing the cache as empty, (2) C-Out: using ID training data with random cropping and initializing the cache with high-entropy samples, (3) D-Out: initializing the cache with OOD samples from OpenImage dataset(different from the test OOD data), and (4) T-Out: initializing the cache with test OOD samples. For each initialization strategy, 800 samples are pre-stored in the cache in advance. In addition, we further explore how different initialization strategies influence the stability during the early stages of testing. We conduct experiments using ImageNet-1K as the ID dataset and Places as the OOD dataset. This experiment is repeated five times, and the AUROC is recorded over the first 10 batch, which are presented in Fig.~\ref{fig:initial_places}. The results show that all initialization methods exhibit unstable performance during the early stages of testing, with the empty initialization being the most severe, but the performance gradually stabilizes as testing progresses. Although initializing the cache with OOD data can partially alleviate early-stage instability, and T-Out even enhances the final OOD detection performance, obtaining OOD samples requires substantial human and material resources. However, in real-world scenarios, the OOD distribution is inherently unknown, making it difficult to acquire samples that match the true distribution. As shown by the results of D-Out, when the initial samples do not reflect the true OOD distribution, the final performance may even be inferior to that of an empty-cache initialization. The C-Out strategy may introduce ID samples into the cache, which can degrade OOD detection performance.

\noindent\textbf{Impact of Cache Update Strategies.} We consider three cache update strategies: (1) first-in-first-out (FIFO), (2) replacing the sample with the highest entropy in the cache (RH), and (3) replacing the sample with the lowest entropy in the cache (RL). Considering that the distribution of test data may shift over time in real-world scenarios, we simulate a temporal distribution shift where OOD data appears sequentially: Textures in $t_0\rightarrow t_1$, iNaturalist in $t_1\rightarrow t_2$, SUN in $t_2\rightarrow t_3$, and Places in $t_3\rightarrow t_4$, denoted as $T\rightarrow i\rightarrow S \rightarrow P$. The OOD detection performance of the three cache update strategies under this setting is shown in Fig.~\ref{fig:update}. During the experiments, both RH and RL strategies tend to retain certain low- or high-entropy samples in the cache for extended periods, making them less likely to be replaced. This reduces the overall data turnover within the cache. In contrast, the FIFO strategy updates solely based on temporal order, ensuring higher cache fluidity and enabling better adaptation to the most recent distribution shifts.

\noindent\textbf{Impact of Cache Construction Strategies.}
DCAC maintains a fixed-capacity cache for each class, making it class-aware. To assess its effectiveness, we compare it against a class-agnostic cache (non–class-aware, NCA), with both caches updated using a FIFO strategy. We simulate a temporal distribution shift in which OOD data appears sequentially, following the ImageNet-1K far-OOD benchmark setup in Fig.\ref{fig:update}. We set DCAC to store up to 20 entries per ID class, and the NCA cache to hold maximum 20000 entries in total. As shown in Fig.~\ref{fig:construction}, DCAC consistently outperforms the NCA cache under these conditions. We attribute this to class bias in the OOD datasets.
For example, when the OOD test set is Places, which contains a large number of landscape images, these samples are more likely to be predicted as ID classes related to scenery. Consequently, the NCA cache tends to accumulate a large number of landscape images. In contrast, DCAC, due to its per-class cache capacity constraint, only stores a limited number of such images.
When the OOD distribution shifts abruptly (e.g., from Places to iNaturalist), the large volume of landscape samples in the NCA cache becomes mismatched with the new distribution, impairing the calibration of incoming data. By maintaining a balanced per-class cache, We substantially reduces the detrimental impact of the previous distribution on the new one.

\noindent\textbf{Hyperparameter Sensitivity Analysis.}
 We evaluate our method through sensitivity experiments. The  hyperparameters evaluated and their ranges are as follows: per-class cache size in the range of [1, 30], calibration strength $\alpha$ in [0.1, 2.5], percentile thresholds $\beta$ for sample selection in [80, 99], and the number of top confidence scores retained per sample $k$ in [1, 1000]. As shown in Fig.~\ref{fig:sensitivity}, our method maintains superior performance over the baseline across wide hyperparameter intervals.

\noindent\textbf{Effectiveness of Calibration.} As shown in Fig.~\ref{fig:score_distribution}, after applying DCAC for prediction calibration, the overlap between ID and OOD samples in MCM score distribution is significantly reduced, further demonstrating calibration efficacy.

\noindent\textbf{Computation Cost.} During inference, DCAC incurs minimal computational overhead. When integrated into CLIP-B/16, it adds 134 MB of memory usage and slightly reduces inference speed from 523 FPS to 511 FPS. For ResNet-50, the memory overhead is 75 MB, with FPS dropping from 1034 to 1027. These results confirm DCAC maintains efficient inference with only modest resource cost. 
% During inference, DCAC incurs minimal time cost. When integrated into CLIP-B/16, it slightly reduces inference speed from 522.64 FPS to 510.74 FPS. For ResNet-50, the FPS drops from 1033.80 to 1026.61. These results confirm that DCAC maintains efficient inference with only a modest time cost.
\section{Further Discussion}

\begin{figure}[t]
    \centering
    \includegraphics[width=\linewidth]{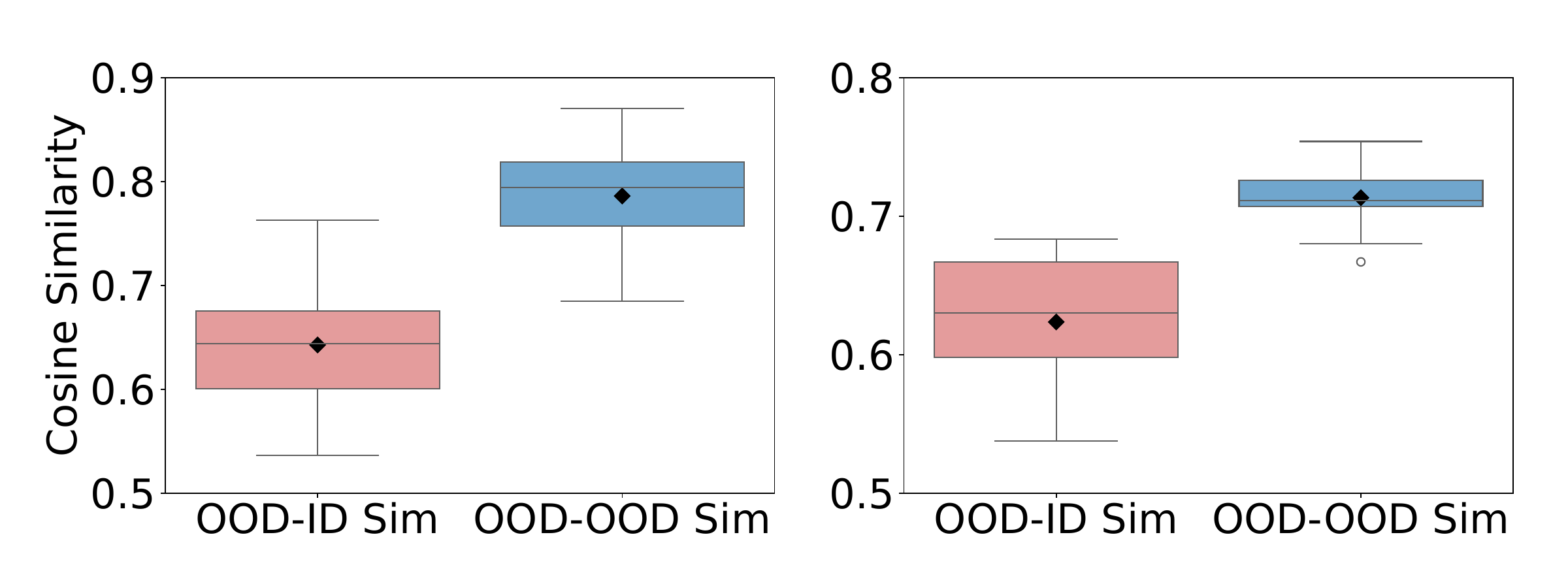}
    \caption{Visual similarity comparison between ID and OOD samples using CLIP-B/16 (left) and ResNet-50 (right). 
    }
    \label{fig:statistics}
\end{figure}

To validate our observation, we compute and analyze the average visual similarity on both ResNet50 and CLIP-B/16 between unconfident OOD samples predicted as each ID class and two groups: (1) overconfident OOD samples assigned high confidence to the same class, and (2) ID samples predicted to be the same class. As shown in Fig.~\ref{fig:statistics}, the average similarity between the two types of OOD samples is higher than that between unconfident OOD samples and ID samples, which verifies our observation.

This phenomenon may be attributed to the following reasons. Previous studies~\cite{ming2022impact,pezeshki2021gradient} have shown that DNNs trained on ID data often overfits background cues shared among samples of the same class, leading to misclassification of OOD samples possessing similar backgrounds. Meanwhile, in VLMs (e.g., CLIP), we observed a distinct issue: some OOD samples, despite clear visual dissimilarities, exhibit unexpectedly high similarity with textual embeddings of specific ID classes, resulting in overconfident predictions. DCAC addresses both scenarios effectively by storing samples with representative background information and misclassified OOD examples and it significantly reduces false confidence without the need for extensive fine-tuning or additional labeled data. %The visualization of the cached samples can be found in Appendix~\ref{further_analysis}.

\section{Conclusion}
This paper tackles the critical problem of overconfident OOD predictions by revealing a key insight: OOD samples with high probability in the same class exhibit greater visual similarity to each other than to true ID samples. Based on this observation, we propose DCAC, a training-free test-time calibration module that maintains class-aware caches to collect unconfident samples and calibrate overconfident predictions. DCAC demonstrates remarkable versatility, seamlessly integrating with diverse OOD detection methods across both unimodal and vision-language models while consistently delivering substantial performance improvements. Our extensive experiments show significant gains on challenging benchmarks with minimal computational overhead. This work provides a practical and effective solution for enhancing model robustness in real-world deployments where distribution shifts are inevitable.
\section*{Acknowledgements}
This work was supported in part by the National Natural Science Foundation of China (Grant No. 62571559), the Major Key Project of PCL (Grant No. PCL2025AS209), the Guangdong Excellent Youth Team Program (Grant No. 2023B1515040025), the Research Start-up Fund for Introduced Talents at Sun Yat-sen University (Grant No. 67000-12255004), and the Young Scientist Program of Sun Yat-sen University (Grant No. 67000-13130003).

\bibliography{aaai2026}

\end{document}